\documentclass{article}

\usepackage[final]{bdl_2018}

\usepackage[utf8]{inputenc} % allow utf-8 input
\usepackage[T1]{fontenc}    % use 8-bit T1 fonts
\usepackage{hyperref}       % hyperlinks
\usepackage{url}            % simple URL typesetting
\usepackage{booktabs}       % professional-quality tables
\usepackage{amsfonts}       % blackboard math symbols
\usepackage{nicefrac}       % compact symbols for 1/2, etc.
\usepackage{microtype}      % microtypography

\usepackage{xcolor}
\usepackage{graphicx}
\usepackage{subcaption} % To enable subfigures
\usepackage{booktabs} % Nice table formatting
\usepackage{csvsimple} % Displaying csv files as tables

\usepackage{caption} 
\hypersetup{
    colorlinks,
    citecolor=black,
    filecolor=black,
    linkcolor=black,
    urlcolor=black
}

\usepackage{amsmath,amssymb,amsfonts,amsthm}
\usepackage{bm}

\usepackage[inline]{enumitem}

\newcommand{\bw}{\bm{w}}
\newcommand{\bz}{\bm{z}}
\newcommand{\bx}{\bm{x}}

\DeclareMathOperator*{\argmax}{argmax}

\title{Bayesian Adversarial Spheres: Bayesian Inference and Adversarial Examples in a Noiseless Setting}

\author{
Artur Bekasov \\ 
University of Edinburgh \\ 
\href{mailto:artur.bekasov@ed.ac.uk}{\ttfamily \fontseries{l}\selectfont artur.bekasov@ed.ac.uk}
\And 
Iain Murray \\ 
University of Edinburgh \\ 
\href{mailto:i.murray@ed.ac.uk}{\ttfamily \fontseries{l}\selectfont i.murray@ed.ac.uk}
}

\begin{document}

\maketitle

\vspace{-8mm}

\section{Introduction}

Modern deep neural network models suffer from \emph{adversarial examples}, i.e.\ confidently misclassified points in the input space \citep{szegedy2014}. Recently \citet{gilmer2018} introduced \emph{adversarial spheres}, a toy set-up that simplifies both practical and theoretical analysis of the problem.

It has been shown that Bayesian neural networks are a promising approach for detecting\footnote{We note that \emph{detecting} is not equivalent to \emph{fixing}. Ideally, we would like our models to classify all in-sample points confidently and correctly. It has been suggested that this might only be achievable by modeling all the invariances present in the data \citep{gal2018}.} adversarial points \citep{rawat2017,li2017,bradshaw2017,gal2018}. Bayesian methods explicitly capture the \emph{epistemic} (or \emph{model}) uncertainty, which we hope will detect parts of the input space that are not covered by training data well enough to justify confident predictions.

In this work, we use the adversarial sphere set-up to understand the properties of approximate Bayesian inference methods in a noiseless setting, where the only relevant type of uncertainty is epistemic uncertainty. We compare predictions of Bayesian and non-Bayesian methods, showcasing the strength of Bayesian methods, although revealing open challenges for deep learning applications. 

\textbf{Contribution} \quad Following our experiments we highlight the following observations:
\begin{enumerate*}
    \item Even a linear model suffers from adversarial examples in the adversarial sphere setup, while careful regularization proves unhelpful.
    \item An accurate Bayesian method (MCMC) makes the model uncertain for adversarial examples, while keeping it reasonably confident for validation points.
    \item The setup presents an example where model uncertainty estimated with bootstrap ensembling is insufficient. 
    \item MCMC results could be improved by using a more flexible prior that enables the model to exploit the symmetry in the problem.
    \item A cheaper variational approximation does not result in an accurate posterior approximation, but demonstrates surprisingly good results on the benchmark. Using a richer variational family does not necessarily result in improved performance down-stream.
\end{enumerate*}

\section{Adversarial spheres}

The adversarial sphere dataset is defined as two concentric hyperspheres with different radii. Each sphere constitutes a manifold for one of the two classes, with points distributed uniformly on the surface. The goal is to learn a decision boundary that would separate the two classes, which we know is itself a hypersphere with a certain radius.

In the original paper, \citeauthor{gilmer2018} show that in high dimensions ($D>60$) we can optimize to find points \emph{on one of the spheres} that the model confidently misclassifies, even if the model demonstrates 100\% accuracy on a huge validation set. Visualisations presented in the paper hint at local overfitting as an explanation for such behavior, which motivates the use of Bayesian methods.

We note that the labels in the dataset are deterministic, i.e.\ there is no inherent \emph{alleatoric} uncertainty in the data, given the correct model. Such setup is interesting because a typical motivation for regularization methods is to avoid fitting the noise in the data. When the problem is noiseless, overfitting could be caused by uncaptured epistemic uncertainty, which is not addressed by standard regularization methods. At the same time, lack of alleatoric uncertainty is inherent to many real-world problems, such as natural image classification.

\section{Bayesian Logistic Regression}

The target decision boundary in the adversarial sphere problem is non-linear. However, it can be represented using logistic regression applied to squared features: 
\begin{gather*}
    P\left(y\!=\!1\,|\,\bm{x},\bm{w}\right) = \sigma\left(\bm{w}^\top\boldsymbol{\phi}(\bm{x})\right),
    \qquad \boldsymbol{\phi}(\bm{x}) = [x_1^2~~\dots~~x_D^2]^\top .
\end{gather*}

This model is able to learn axis-aligned ellipsoidal decision boundaries in $D$ dimensions. Learned NN basis could also be used, as explored by \citet{snoek2015}, where Bayesian logistic regression could be framed as being Bayesian about the last layer of a neural network. Inference gets more difficult for all parameters of a neural network, hence we must convince ourselves that we fully understand properties of inference methods for this simpler model.  

Choosing a prior for Bayesian logistic regression, especially in a linearly separable setting, is not trivial \citep{gelman2008}. In such setting we are seeking large weights values, in order to make confident predictions. This implies the use of broad, uninformative (or weakly informative) priors. (In our experiments we use an isotropic Gaussian with large width $\sigma_w = 100$.)

\section{Experiments}

\begin{figure}[t]
\centering
\begin{subfigure}{.325\textwidth}
    \centering
    \includegraphics[width=\textwidth]{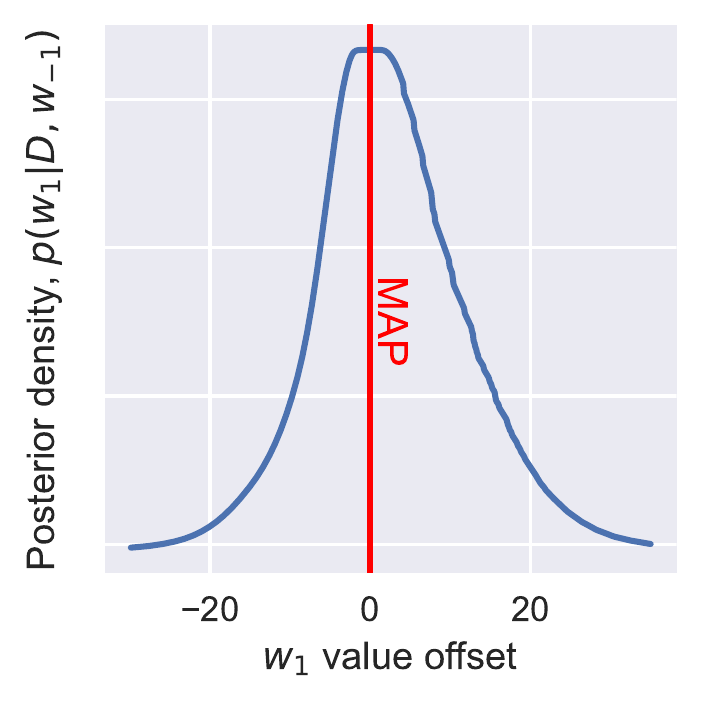}
\end{subfigure}
\begin{subfigure}{.325\textwidth}
    \centering
    \includegraphics[width=\textwidth]{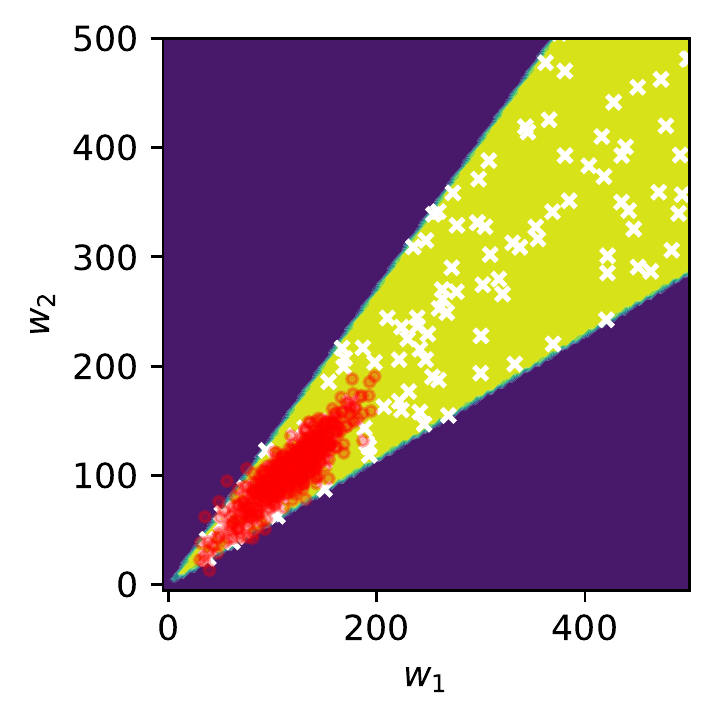}
\end{subfigure}
\begin{subfigure}{.325\textwidth}
    \includegraphics[width=\textwidth]{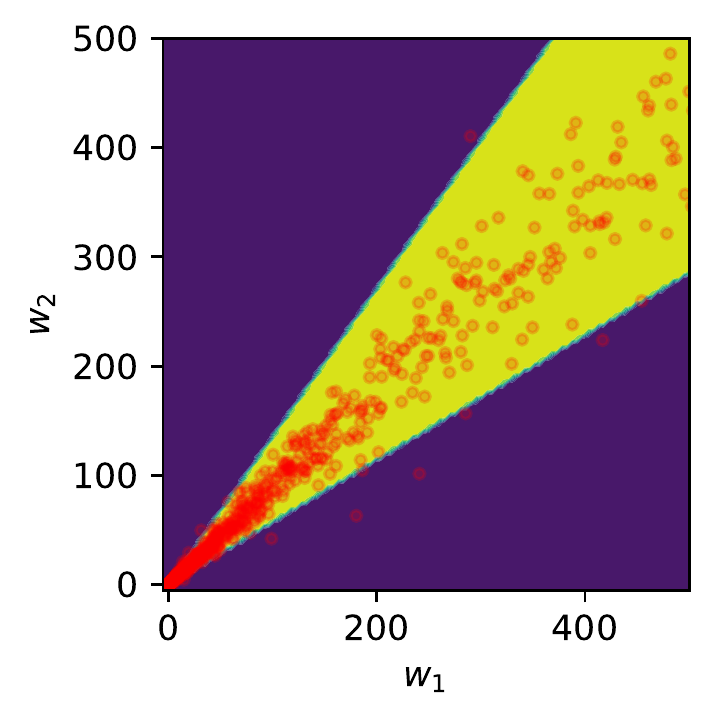}
\end{subfigure}
    \caption{ \small
        \textbf{(left)} Posterior density for one of the weights in a 500-dimensional setting, conditioned on the MAP values for all the other weights. The overall shape is Gaussian, but the region around the mode is flat. This is due to a broad prior used, and linear separability of the data.
        \textbf{(middle)} Background: likelihood for a 2D sphere problem. Red circles: samples from a variational posterior. White crosses: MCMC samples. Variational posterior captures the uncertainty about the ``direction'' in the weight space, but rules out large weights that are typical under the true posterior.
        \textbf{(right)} Background: likelihood for a 2D sphere problem. Red circles: samples from a variational posterior for a \emph{hierarchical parameterization}. While most of the mass is still near the origin, higher weight values no longer have zero density. 
    }
    \label{fig:fig}
\vspace{-4mm}
\end{figure}

\begin{table}[t]
\small
\centering
\begin{filecontents*}{fig/results.csv}
MAP,1.000,0.999,--
Laplace,\textcolor{red}{0.501},\bftab{0.499},--
Bootstrap,1.000,0.961,0.957
MCMC,0.976,\bftab{0.558},0.205
SVI (MC),0.991,0.606,0.516
Hier. SVI (MC),0.978,0.678,0.561
MCMC (non-zero mean),1.000,\bftab{0.341},0.301
\end{filecontents*}
\csvreader[no head, 
tabular=lrrr,
table head={\toprule Model & Avg.\ confidence $\uparrow$ & Adv.\ err. $\downarrow$ & Resampled err. $\downarrow$ \\ \midrule},
table foot=\bottomrule,
late after line=\ifboolexpr{
    test {\ifnumequal{\thecsvrow}{4}} 
    or 
    test {\ifnumequal{\thecsvrow}{6}}
    or 
    test {\ifnumequal{\thecsvrow}{7}}
}
{\\\midrule}{\\},
late after last line=\\
]
{fig/results.csv}
{1=\model,2=\pred,3=\adv,4=\secondens}
{\model & \pred & \adv & \secondens}
    \caption{ \small
        Avg.\ confidence as measured on the validation set: ${\mathop{\mathbb{E}}_{x \sim X_{\mathit{val}}}\left[p\left(y_{true} \mid x\right)\right]}$. Adversarial error is defined as ${\mathrm{max}_x p(y=\left(1 - y_{\mathit{true}}\right) \mid x)}$. Resampled error is calculated by sampling a second ensemble and calculating its prediction on the adversarial points found for the original ensemble. Arrows indicate whether we are looking for higher or lower values. 1000 samples/models used for all ensembles. } 
\label{tab:results}
\vspace{-7mm}
\end{table}

Results for a 500-dimensional adversarial sphere dataset are summarized in Table~\ref{tab:results}.

\textbf{Maximum Likelihood / MAP} \quad 
We first train a logistic regression model using maximum likelihood. For a large training set the model becomes immune to adversarial attacks, but on a smaller training dataset (with 1000 datapoints) the model demonstrates perfect accuracy on the validation set (error rate below $10^{-5}$), but at the same time we can find points which the model misclassifies with more than $99\%$ confidence. 
Switching to simple penalized ML / MAP with a Gaussian prior on the weights (i.e.\ adding L2 regularization) does not resolve the issue. We note the following contradiction: we know that in the ``true'' solution weight magnitudes are large, as the dataset is linearly separable, yet we are penalizing such values.

\textbf{MCMC sampling} \quad
We then attempt to understand what an accurate Bayesian method \citep[slice sampling,][]{neal2003} would do.
The results show that these approximate Bayesian predictions become uncertain for adversarial points. While we also get reasonably confident predictions on points from the validation set, the confidence is reduced when compared to ML/MAP results. Not all validation samples are close to the (limited) training data, hence the model cannot be completely confident in its predictions without exploiting symmetries/invariances in the data.

\textbf{Bootstrap} \quad
Can we get similar results from a simpler method? We train an ensemble of models using bootstrap sampling, and use an averaged prediction during test time. Bootstrap ensembles are often claimed to be an approximation to the Bayesian posterior \citep[§8.4]{hastie2001}. 
In our setup, however, the uncertainty estimated in this way is insufficient --- the worst adversarial error is hardly reduced. Moreover, the rightmost column of Table~\ref{tab:results} shows that the adversarial points found are transferable to other ensembles trained in the same way. In other words, the adversarial procedure learns to exploit the training method, and not a particular ensemble, which is not the case for MCMC.

\textbf{Laplace approximation} \quad
Sampling, while accurate, is often impractical. We look into a cheaper method of Laplace approximation, that fits a Gaussian to the posterior by matching the curvature at the mode \citep[Chapter 27]{mackay2003}.% 
We observe that while the method also detects adversarial examples, it becomes just as uncertain on validation points. The issue is caused by the non-Gaussian shape of the posterior near the mode, as illustrated in Figure~\ref{fig:fig}, which stems from multiple steep decision boundaries having the same likelihood given the data. This results in an unrealistically wide Gaussian approximation and uncertain predictions for all points. This phenomenon is also discussed by \citet{kuss2005}.

\textbf{Variational approximation} \quad
Variational inference is an alternative way of fitting a simple distribution family to the true posterior. We implement and evaluate Stochastic Variational Inference \citep[SVI,][]{hoffman2013,ranganath2013} with a full-covariance Gaussian family in our experiments.
Variational inference makes probabilistic predictions that are surprisingly close to the ones of MCMC. At the same time, when we look at the samples from the true posterior, as shown in Figure~\ref{fig:fig}, the fit is clearly not perfect. In particular, the variational posterior assigns near-zero density to higher weight values, ruling out the steepest decision boundaries, which are closest to the truth.

\textbf{Hierarchical model} \quad
An alternative model could be used to enable more accurate variational approximation. We reparameterize the model using a hierarchical approach: $w_i \sim N(0, e^v); v \sim N(0, \sigma_{v}^2)$. Equivalently, using a ``non-centered'' parameterization: $w_i = e^{v/2} z_i; z_i \sim N(0, 1); v \sim N(0, \sigma_{v}^2)$. We can then fit two variational distributions, $q_{\bz}(\bz)$ and $q_v(v)$. Intuitively, we are defining a distribution over the ``direction'' in the weight space using $\bz$, and the positive ``distance'' in that direction using $v$. This is related to the work by \citet{ranganath2016}, but where we also update the prior to match the variational family.
Figure~\ref{fig:fig} shows that such parameterization results in a more sensible posterior fit, where we no longer assign zero density to larger weight values. However, in this case the down-stream performance is not improved. 

\textbf{Exploiting symmetry} \quad
Similarly, we can use a hierarchical parameterization to allow the model to exploit the symmetry in the problem. We know that the max margin decision boundary is a sphere, hence the ``true'' weight values must be close to one another. We thus assume the weights come from a prior distribution $w_i \sim N(\mu, \sigma_w^2)$, where the mean has its own hyper-prior $\mu \sim N(0, \sigma_{\mu}^2)$. 
This results in a model with well calibrated uncertainty, as seen in the last row of Table~\ref{tab:results}. Some of the earliest work on Bayesian neural networks recognized the importance of choosing hierarchical priors carefully \citep{neal1994}. The prior we use here is favored by the data, but trying it was guided by our knowledge of the problem. It is likely that in realistic problems with deeper networks, the choice of prior will only have a stronger effect on the uncertainties reported by Bayesian methods. Exploring the families of priors that can capture symmetries and invariances in real problems is an important direction in Bayesian deep learning.  

\clearpage

\section*{Acknowledgments}

This work was supported in part by the EPSRC Centre for Doctoral Training in Data Science, funded by the UK Engineering and Physical Sciences Research Council (grant EP/L016427/1) and the University of Edinburgh. Authors thank James Ritchie for his proof-of-concept implementation for this work.

\bibliographystyle{hapalike}
\bibliography{/Users/artur/git/papers/bibliography}

\appendix

\section{Details of training} 

We use 1000 training samples and 100k validation samples in a 500-dimensional setting for our experiments.

We normalize the input features of the logistic regression after applying the basis functions, as our preliminary experiments have revealed that normalization plays an important role in the performance of some methods.

We use the Pytorch implementation of the LBFGS optimizer for MAP, bootstrap and Laplace experiments.\footnote{Note that this version of LBFGS does not implement line search.}  SGD with momentum is used for variational methods, with batch size of 100, learning rate of 0.01 and momentum coefficient of 0.98. Limited hyperparameter exploration was performed. We run optimization for 50 thousand iterations. 

A spherical Gaussian prior with $\sigma_{w} = 100$ is used for experiments with a standard model. Given feature normalization, this represents a reasonably broad belief. For the hierarchical model, we use $\sigma_{v} = 100$.

\section{Details of adversarial optimization}
\label{sec:adv_opt}

The goal of adversarial optimization could be formalized as follows:
\begin{gather*}
\bx_{\mathit{adv}} = \argmax_{\bx} p(y=y_{\mathit{target}}|\bx); \quad y_{\mathit{target}} = 1 - y_{\mathit{true}}
\end{gather*}

In our work, we would also like to restrict $\bx$ to lie on a surface of a sphere with a given radius. In line with \citet{gilmer2018}, we solve the constrained optimization problem using Projected Gradient Descent, which works by projecting the current point onto the required sphere after every gradient step.

If we attempt to use a gradient based optimizer with such objective, however, we are likely to run into numerical problems. Logistic regression model is defined as
\begin{gather*}
p(y=1|\bx) = \sigma(a),
\end{gather*}
where $\sigma$ is a sigmoid function and $a = \bw^{\top}\bx$. For a model trained with MLE or MAP, a random point $x$ on one of the spheres will typically result in a large magnitude of $a$, as model's predictions are extremely confident. This, in turn, will select a point at one of the tails of $\sigma(a)$. The sigmoid function saturates for relatively small activation magnitudes, especially when using single floating-point precision. This means that $\nabla_a \sigma(a) = 0$, and we will not be able to make any meaningful optimization steps.

To fix this, we use the fact that the sigmoid is a monotonically increasing function, hence:
\begin{gather*}
\argmax_{\bx} p(y=y_{\mathit{target}}|\bx) = \argmax_{\bx} \sigma(a) = \argmax_{\bx} a.
\end{gather*}
In other words, we can optimize the logit of a prediction, rather the prediction itself, hence avoiding numerical issues outlined above.

The same trick is not applicable to ensembles, however. Ensemble prediction is defined as
\begin{gather*}
p(y=1|\bx) = \frac{1}{M} \sum_{i=1}^M \sigma(a_m), 
\end{gather*}
where $a_m$ is an activation of $m$-th model in the ensemble. Intuitively, we can not maximize this objective by maximizing the mean of activations. The mean of activations could be maximized by pushing one of the activations to infinity. This would clearly not maximize the original objective, however, as saturation of the sigmoid limits the contribution of each model's prediction.

Instead, we use another trick. We know that log probabilities have better numerical properties than probabilities themselves, and that log is also a monotonically increasing function. Rewriting the objective in terms of log probabilities we get
\begin{gather*}
\log p(y=1|\bx) = \log \frac{1}{M} \sum_{i=1}^M \sigma(a_m) = \log \sum_{i=1}^M \sigma(a_m) + const. 
\end{gather*}
Then, we can apply Jensen's inequality to set a lower bound on the log sum:
\begin{gather*}
\log \sum_{i=1}^M \sigma(a_m) \ge \sum_{i=1}^M \log \sigma(a_m),
\end{gather*}
where $\log \sigma(a_m)$ could be expressed as $-\text{softplus}(-a_m)$, which has a numerically stable implementation. We could then optimize this lower bound instead of the original objective, at least to guide the optimization to a more numerically stable region during early steps. 

We note that it has also been proposed to use outlined numerical issues as defense mechanism, to make it difficult for an attacker to obtain meaningful gradients for adversarial optimization. Recent work by \citet{athalye2018} discusses these issues further, and outlines various ways of defeating such defense in different models.

In our experiments, we optimize the more numerically stable lower-bound of the loss for 300 iterations before switching to the real optimization criterion. Step size of 0.01 is used for most experiments, but we lower it to 0.0001 for Monte-Carlo SVI and sampling experiments, due to observed instability during optimization. Optimization is terminated if the best achieved loss has not noticeably improved for 10 iterations. We consider an absolute difference in loss value of more than $1\mathrm{e}{-4}$ to be a noticeable improvement. This number was picked empirically, to strike the balance between good convergence and the amount of computation.

\end{document}